\documentclass[11pt,a4paper]{article}
\usepackage[hyperref]{acl2019}
\usepackage{times}
\usepackage{latexsym}

\usepackage{url}
\usepackage{multirow}
\usepackage{amsfonts}
\usepackage{graphicx} 
\usepackage{amsmath}
\usepackage{mathtools}
\usepackage[ruled,vlined]{algorithm2e}
\usepackage{fancyhdr,amssymb}

\usepackage[utf8]{inputenc}

\usepackage{todonotes}

\aclfinalcopy 

\title{What Matters for Neural Cross-Lingual Named Entity Recognition: \\An Empirical Analysis}

\author{Xiaolei Huang$^{1*}$, 
Jonathan May$^2$, 
Nanyun Peng$^2$\\
\textsuperscript{1}Dept. of Information Science, University of Colorado Boulder\\
\textsuperscript{2}University of Southern California Information Sciences Institute \\
xiaolei.huang@colorado.edu, \{jonmay, npeng\}@isi.edu
\thanks{~~The work was done when the first author worked as an intern at USC ISI.}
}

\date{}

\begin{document}
\maketitle
\begin{abstract}
Building named entity recognition (NER) models for languages that do not have much training data is a challenging task. While recent work has shown promising results on cross-lingual transfer from high-resource languages to low-resource languages, it is unclear what knowledge is transferred. In this paper, we first propose a simple and efficient neural architecture for cross-lingual NER. Experiments show that our model achieves competitive performance with the state-of-the-art. We further analyze how transfer learning works for cross-lingual NER on two transferable factors: sequential order and multilingual embeddings, and investigate how model performance varies across entity lengths. Finally, we conduct a case-study on a non-Latin language, Bengali, which suggests that leveraging knowledge from Wikipedia will be a promising direction to further improve the model performances. Our results can shed light on future research for improving cross-lingual NER.

\end{abstract}

\section{Introduction}
Named Entity Recognition (NER) is an important NLP task that identifies the boundary and type of named entities (e.g., person, organization, location) in texts. However, for some languages, it is hard to obtain enough labeled data to build a fully supervised learning model.
Cross-lingual transfer models, which train on high-resource languages and transfer to target languages are a promising direction for languages with little annotated data~\cite{bharadwaj2016phonologically, tsai2016cross, pan2017cross, yang2017transfer, mayhew2017cheap, wang2017multi, cotterell2017low, feng2018improving, xie2018neural, zhou2019dual}.

Cross-lingual NER models learn transferable knowledge mainly from four sources: translation~\cite{mayhew2017cheap, feng2018improving}, bilingual embeddings~\cite{ni2017weakly, xie2018neural}, phonetic alphabets~\cite{bharadwaj2016phonologically} and multi-source learning~\cite{mayhew2017cheap, lin2018multi, zhou2019dual, chen2019multi, rahimi2019multilingual}. Translation can be applied at either the sentence level via machine translation or at the word and phrase level via application of bilingual dictionaries. Bilingual embeddings are a form of bilingual dictionaries; they constitute a projection from one pre-trained language representation into the same vector space as the other such that words with the same meaning have similar representations~\cite{conneau2017word}. Phonetic alphabets enable different languages to share the same pronunciation characters so that the character-level knowledge is transferable between languages that otherwise have different character sets, such as English and Bengali~\cite{hermjakob2018out}. Multi-source learning is effective when multiple or similar language resources are available by learning shareable knowledge. For example, training a model on both English and Hindi can significantly improve the model performance on Bengali than only using English~\cite{mayhew2017cheap}. However, there is little prior work with detailed analysis of how cross-lingual NER models transfer knowledge between languages on different levels.

In this paper, we focus on a single-source zero-shot transfer setting where we transfer from English to target languages that have no annotated data. In our settings, the resources are limited to annotated source language data, bilingual dictionaries, and unlabeled corpora from both source and target languages. We first propose a neural cross-lingual NER model that combines the ideas of translation, bilingual embedding, and phonetic alphabets. Next, we conduct qualitative analyses to answer the following questions on how the model transfers knowledge under the cross-lingual settings: 1) does the source language syntax matter? 2) how do word and character embeddings affect the model transfer? We analyze how F1 scores differ across different entity lengths. Finally, we conduct a case study on Bengali.

\begin{figure*}[htp]
\centering
\includegraphics[scale=0.53]{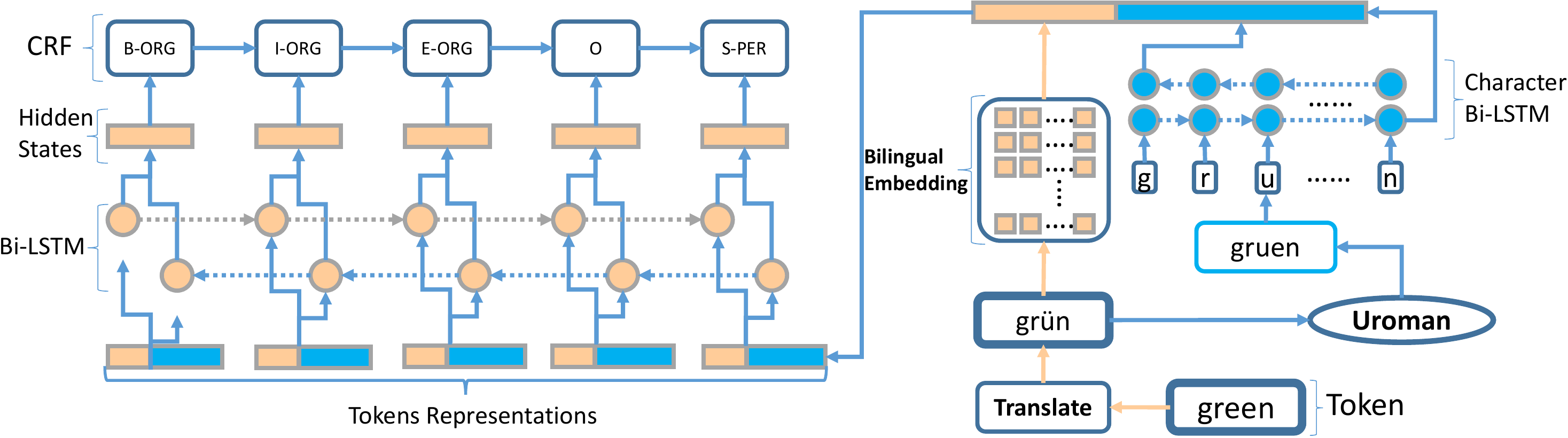}
\caption{Architecture of our proposed model. Each word is translated from the source. Each token representation contains two concatenated parts: a bilingual word embedding (light orange) and character representations (blue). The character representation for each word is generated via the Bi-LSTM. Words are first transliterated and split into individual characters before being fed into the character Bi-LSTM.}
\label{fig:model}
\end{figure*}

\section{Model}
\label{sec:model}

NER models take a sequence of tokens as input and predict a sequence of labels such as person (PER) or location (LOC). In this paper, we adopt the neural architecture NER model from~\newcite{lample2016neural,ma2016end}. The model first combines pre-trained word and character embeddings as token representations, then feeds the representations to one layer of a Bidirectional Long Term Short Memory (Bi-LSTM)~\cite{hochreiter1997long}, and finally predicts the outputs via a Conditional Random Field (CRF). We show the model architecture on left of Figure~\ref{fig:model}. However, languages express the same named entity in different words and characters. To bridge the barriers, we combine three strategies: bilingual embedding, reverse translation, and transliteration.

\textbf{Bilingual Embedding.} Word embeddings are usually trained for each language separately and therefore in different vector spaces~\cite{ruder2019survey}.
To map word embeddings into the same shared space, we use  MUSE~\cite{conneau2017word} to build our bilingual embeddings. The bilingual embeddings enable the words with the same meaning to have similar word representations in the shared space.

\textbf{Reverse Translation.} We use the bilingual dictionaries provided by \newcite{rolston2016collection}. Language variations exist among source and target languages and training on monolingual corpus might limit learning the variations. Therefore, we translate the source language, English, to the target language for the training corpus to reconstruct source language sentences and learn source and target languages jointly. However, one English word might have multiple corresponding translations. To select the best translation, we define an empirical score function $F(w, w_{t, i})$ that calculates the cosine similarity between the translated target word ($w_{t, i}$) and the English source word ($w$) based on its contextual words ($w_{c, j}$).
\begin{multline}
    F(w, w_{t, i}) = \alpha \cdot \cos (E(w), E(w_{t, i}))\\ + (1-\alpha)\cdot\sum_{j=1}^m\frac{\cos(E(w_{t, i}), E(w_{c, j}))}{(d_j+1)^2}
\end{multline}

\noindent where $E(w)$ is the bilingual embedding vector of the word, $d_j$ is the sequential distance between the word and its contextual word j, and  $\alpha$ is the trade-off factor that balances impacts of translation pair and contextual words. In this study, we set $\alpha$ to 0.5. We choose the translation pair with the highest similarity score. Note that the reverse only applies during the training step that translates  English into the target language.

\begin{table}[htp]
\centering
\begin{tabular}{c|c|c|c}
Target word & Source word & Uroman & Type \\\hline
grünen & green & gruenen & ORG \\ \hline
\includegraphics[scale=0.04]{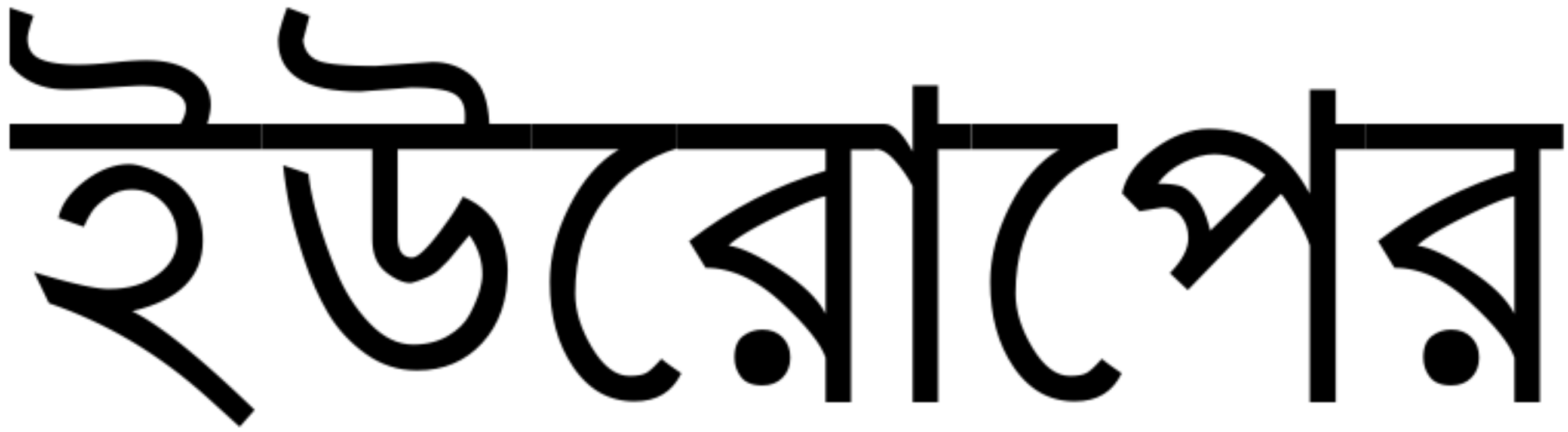} & Europe & iuropera & LOC \\\hline
\end{tabular}
\caption{Examples of Uroman that maps different languages into the same character space. We show two target languages, German (top) and Bengali (bottom). The second column is a translation of the target language. The transliterations show the phonetic similarity between the source and target languages.}
\label{tab:uroman}
\end{table}

\textbf{Character - Uromanization.} Different languages may not share the same characters, but some named entities in the multilingual corpora share phonetic similarities~\cite{huang2004improving}. To map multilingual corpora into the same character space and connect the phonetic similarity between entities, we employ Uroman\footnote{\url{https://www.isi.edu/~ulf/uroman.html}}~\cite{hermjakob2018out}, which transliterates any language into something roughly pronounceable English spellings, while keeps the English words unchanged. We show some examples of Uroman in Table~\ref{tab:uroman}.

\section{Experiments}
In this study, we first evaluate our proposed cross-lingual method on the CoNLL 2002 and 2003 NER datasets~\cite{TjongKimSang:2002:ICS:1118853.1118877, tjong2003introduction}. We then conduct ablation studies to examine how the model learns transferable knowledge under the cross-lingual settings. Finally, we conduct a case study on Bengali, a low resource language.

\subsection{Data}


The CoNLL datasets contain four different European languages, Spanish, Dutch, English, and German. The data contains four types of named entities: person (PER), organization (ORG), location (LOC) and MISC. We use Bengali language data from LDC2015E13 (V2.1). To be consistent, we only keep the PER, ORG and LOC tags and ignore MISC tags from the Bengali corpus.

In this study, we choose the BIOSE tag schema instead of standard BIO, where the B, I, O, S, E refer to the beginning, inside, outside, single and end of an entity, respectively. Previous work shows that BIOSE can learn a more expressive model than the BIO schema does~\cite{ratinov2009design}. We lowercase the data and replace numbers and URLs as ``num'' and ``url'' respectively. We use the training data to train our model, the development sets to select the best well-trained model, and the test sets to evaluate performance, where the training data is English and the development and test sets are from the target language. 

\subsection{Experimental Settings}
\textbf{Bilingual Embedding}. We use the 300-dimensional bilingual word embeddings pre-trained by MUSE~\cite{conneau2017word} and 300-dimensional randomly initialized character embeddings. Specifically, we first collect monolingual pre-trained word embeddings from  fastText~\cite{bojanowski2016enriching}. We then align the embeddings using the MUSE supervised model. We normalize the inputs to MUSE and keep the other hyperparameters as defaults. We merge the aligned word embeddings and remove duplicates. Updating the embedding during training time will change the vector space of partial bilingual embeddings and therefore break the vector alignment. Thus, we freeze the embedding weights during the training step. For OOV (out-of-vocabulary) words, we use a randomly initialized 300 dimension vector within $[-0.1, 0.1]$.

\textbf{Reverse Translation}. We follow the general translation step in Section~\ref{sec:model}. Specifically, we replace English words with target language words if the translation pairs exist in the bilingual dictionaries~\cite{rolston2016collection}. If a translation pair does not exist for a word, we keep the word unchanged in the training data. We use the pre-trained bilingual embedding to help select the best choice of polysemy.

\textbf{Model parameters}. We use 300-dimension hidden states for both character and token level Bi-LSTMs. To prevent overfitting, we apply dropout~\cite{srivastava2014dropout} with a rate of 0.5 on outputs of the two Bi-LSTMs. We follow the Conditional Random Field (CRF) setup of~\newcite{peng2015named}. We then randomly initialize the weights of the layers within $[-0.1, 0.1]$. We train the model for 200 epochs and optimize the parameters by Stochastic Gradient Descent (SGD) with momentum, gradient clipping, and learning rate decay. We set the learning rate (lr) and the decay rate (dr) as 0.01 and 0.05 respectively. We update the learning rate by $\frac{lr}{(1 + (n+1)*dr)}$ after epoch $n$. We clip the gradients to the range of [-5.0, 5.0].
We measure performance by F1 score.

\subsection{Baselines}
In this study, we compare our proposed method with three close works under the cross-lingual settings. We compare our method with the best-reported performance from their works. We briefly summarize the the three baselines in this section.
\begin{itemize}
\setlength\itemsep{.1em}
    \item WikiNER~\cite{tsai2016cross} first links named entities in the multilingual Wikipedia corpora and extracts page categories as ``wikifier'' features, and use these features to achieve cross-lingual transfer.
    \item CTNER~\cite{mayhew2017cheap} first translates the corpora into English via bilingual dictionaries and multilingual Wikipedia entries on both word and phrase levels, and directly perform NER on the translated target language.
    \item NCNER~\cite{xie2018neural} proposes a neural NER model with attention mechanism. The work is closest to ours. However, our model takes different approaches in obtaining multi-lingual embedding, translation, and transliteration as ~\newcite{xie2018neural}, while the neural models share the similar architecture of Bi-LSTMs-CRF, and use multi-lingual embeddings to achieve transfer. 
\end{itemize}

\subsection{Results}

\begin{table}[htp]
\centering
\begin{tabular}{c|c|c|c}
Models & Spanish & Dutch & German \\\hline
WikiNER* & 60.55 & 61.60 & 48.10 \\
CTNER* & 51.82 & 53.94 & 50.96 \\
CTNER*+ & 65.95 & 66.50 & 59.11 \\
NCNER & \textbf{72.37} & 71.25 & 57.76 \\\hline
Our Model & 64.48 & \textbf{73.44} & \textbf{62.26}
\end{tabular}
\caption{F1 score comparisons of cross-lingual models on Spanish, Dutch and German. The ``*'' indicates the model uses the Wikipedia resources. The ``+'' means training model by multiple language resources.}
\label{tab:exp}
\end{table}

As seen in Table~\ref{tab:exp}, our proposed method outperforms the previous works in nearly all cross-lingual tasks. Moreover, the two neural model methods exceed the performance of the other models. These results suggest the effectiveness of neural cross-lingual transfer through multi-lingual embedding space, and show competitive results of our proposed methods.  

\subsection{Ablation Studies}

\begin{table}[htp]
\centering
\begin{tabular}{c|c|c|c}
Models & Spanish & Dutch & German \\\hline
Full Model & 64.48 & 73.44 & 62.26 \\ \hline
Shuffle & 49.44 & 40.61 & 27.25 \\
Word-only & 53.00 & 59.60 & 52.58 \\
Char-only & 18.88 & 12.15 & 16.46 \\
\end{tabular}
\caption{F1 scores of different ablation analyses, compared to our full model.. ``Shuffle'' means the training data is shuffled. ``Word-only'' means our proposed model is only fed bilingual word embeddings, and ``Char-only'' means the model only receives character embeddings  as input.}
\label{tab:ablation}
\end{table}

We conduct ablation analyses to understand how the model transfers learned knowledge from one language to the other. We focus on two aspects: syntax and embeddings.

\textbf{Syntax analysis.} Different languages might not share the same syntax structures. The neural model learns sequential information from the source language and applies the learned knowledge on the target language. The importance of this sequential information is unknown. We shuffle the sentences of training data while keeping the internal order of named entities unchanged. We show the results in Table~\ref{tab:ablation}. The results show that after shuffling, the performances of our model decrease and the decreases vary among different languages. This suggests the importance of sequential information for those languages under the cross-lingual settings.

\textbf{Embeddings analysis.} We train our proposed models with both character and bilingual word embeddings, however, how the model values different embeddings is unknown. We feed the model with either bilingual word or character embeddings. Table~\ref{tab:ablation} shows that bilingual word embeddings have better performance than character embeddings across the three languages. 
The results suggest that the aligned bilingual embeddings are more important than the character embeddings for the three languages.

\subsection{Entity length analysis} 
While we can observe how models perform overall on the named entities, we can not know how models  differ from short to long entities. 
Particularly, probing if models hold strong biases towards shorter entities can help interpret the process of cross-lingual transfer learning. 
To make the comparison, we first categorize entities into three levels: single token, two tokens and greater or equal to three tokens.
We then calculate the F1-score of the three entity lengths in the test set and among correctly predicted entities.
Finally we summarize the F1-score in Table~\ref{tab:f1len}.


\begin{table}[htp]
\centering
\begin{tabular}{c|ccc}
Language & 1 & 2 & $\geq$3 \\\hline
Spanish & 68.93 & 68.21 & 47.57 \\
Dutch & 69.49 & 82.46 & 60.85 \\
German & 59.50 & 70.76 & 37.22
\end{tabular}
\caption{F1 scores of different lengths of entities across the three languages: Spanish, Dutch and German. The number 1 refers to the entities with single token, the 2 means the entities with two tokens and the $\geq$ 3 indicates the entities with not less than three tokens.} 
\label{tab:f1len}
\end{table}

With comparing to the overall performance across the three languages in Table~\ref{tab:exp}, we can observe that the single token shows relatively closer to the performance, the entities with two tokens achieve the higher scores, while the entities with more than 2 tokens decrease significantly ranging from 12.59 to 25.04 absolute percentages of F1 scores.
The observation indicates that entities longer than two tokens are more difficult to infer. 
This might encourage us to balance the weight of long entities in our current evaluation method which ignores entity length when datasets have high volumes of long entities.

\section{Case Study: Bengali}
\begin{table}[htp]
\centering
\begin{tabular}{c|c}
Models & F1 score \\\hline
CTNER & 30.47 \\
CTNER+ & 31.70 \\
CTNER* & 46.28 \\
CTNER*+ & 45.70  \\\hline
Our Model & 34.29 
\end{tabular}
\caption{F1 score comparisons of translation-based models on Bengali. The ``*'' indicates the model uses Wikipedia resources. The ``+'' means a model is trained with multiple language resources.}
\label{tab:bengali}
\end{table}
The previous cross-lingual settings were only for European languages, which share similar alphabets. However, many languages use non-Latin orthography. In this work, we present a case study on Bengali,
which does not use a Latin alphabet. We compare our proposed method with the translation-based method, CTNER~\cite{mayhew2017cheap}. The results in Table~\ref{tab:bengali} show that our model outperforms the previous methods without Wikipedia. 

The results suggest multilingual Wikipedia is critical for future performance improvements beyond simple transfer. This is to be expected; a domain discrepancy exists between the source and target language data and therefore many named entities are missing or mismatched: partial data sources of the Bengali come from social media and online forums~\cite{cieri2016selection}. By contrast, the only data source of CoNLL data is from news articles~\cite{TjongKimSang:2002:ICS:1118853.1118877, tjong2003introduction}. While transfer can help provide universal context clue information across languages there is no substitute for a resource of actual names.

\begin{table}[htp]
\centering
\begin{tabular}{c|cc}
Language & Type & Token \\\hline
Spanish & 2.5 & 0.8 \\
Dutch & 2.5 & 0.9 \\
German & 2.3 & 1.5 \\
Bengali & 18.2 & 12.2
\end{tabular}
\caption{OOV rate (percentage) in our bilingual word embeddings across the four languages. Type indicates unique words, and token refers to counting token numbers.} 
\label{tab:oov}
\end{table}

The colloquium words from social media may cause the issue of out of vocabulary (\textit{OOV}) and further impact the transferring process.
We summarize the percentage of missing words in our pre-trained bilingual word embeddings in Table~\ref{tab:oov}.
The OOV rate of Bengali is significantly higher than that of the other three languages.
This suggests that the high OOV rate and the discrepancy of data sources may hurt the effectiveness of transfer.

\section{Conclusion}
We have presented a simple but efficient method to adapt neural NER for the cross-lingual settings. The proposed method benefits from multiple transferable knowledge and shows competitive performances with the state of the art using limited resources. We examine multiple factors that impact the transfer process and conduct an ablation study to measure their influences. 
Our experiment on Bengali shows that leveraging knowledge from Wikipedia will be a promising direction for future research.

\section{Acknowledgments}
This work is partially funded by DARPA (HR0011-15-C-0115) and NIH R01 (LM012592).
The authors thank the anonymous reviewers for their helpful suggestions. 
We thank Boliang Zhang, Stephen Mayhew, Michael J. Paul and Thamme Gowda for their useful feedback.
Any opinions, findings, conclusions, or recommendations expressed here are those of the authors and do not necessarily reflect the view of the sponsor.

\bibliography{emnlp2019}
\bibliographystyle{acl_natbib}

\end{document}